\pdfoutput=1

\documentclass[11pt]{article}

\usepackage{EMNLP2023}

\usepackage{times}
\usepackage{latexsym}
\usepackage{algorithm}
\usepackage{algpseudocode}
\usepackage{booktabs}
\usepackage{tabularx}

\usepackage[T1]{fontenc}

\usepackage[utf8]{inputenc}

\usepackage{microtype}

\usepackage{inconsolata}

\usepackage{amsmath}
\usepackage{amsfonts}
\usepackage{dsfont}
\usepackage{breqn}
\usepackage{multirow}
\usepackage{graphicx}
\usepackage{subfigure}

\usepackage{colortbl}
\definecolor{Ocean}{RGB}{129,194,234}

\usepackage{color, soul}
\definecolor{tri_red}{RGB}{187,39,26}
\definecolor{tri_blue}{RGB}{75,119,209}
\definecolor{tri_green}{RGB}{120,166,90}
\definecolor{pipeline_red}{RGB}{187,39,26}
\definecolor{pipeline_green}{RGB}{71,116,44}
\definecolor{table_ocean}{RGB}{229,242,250}
\sethlcolor{table_ocean}

\usepackage{cleveref}
\crefformat{section}{\S#2#1#3}
\crefformat{subsection}{\S#2#1#3}
\crefformat{subsubsection}{\S#2#1#3}
\crefrangeformat{section}{\S#3#1#4 to~\S#5#2#6}
\crefmultiformat{section}{\S#2#1#3}{ and~\S#2#1#3}{, #2#1#3}{ and~#2#1#3}
\Crefformat{figure}{#2Fig.~#1#3}
\Crefmultiformat{figure}{Figs.~#2#1#3}{ and~#2#1#3}{, #2#1#3}{ and~#2#1#3}
\Crefformat{table}{#2Tab.~#1#3}
\Crefmultiformat{table}{Tabs.~#2#1#3}{ and~#2#1#3}{, #2#1#3}{ and~#2#1#3}
\Crefformat{appendix}{#2Appx.~\S#1#3}
\crefformat{algorithm}{Alg.~#2#1#3}
\Crefformat{equation}{#2Eq.~#1#3}

\usepackage{tcolorbox}

\newcommand{\stitle}[1]{\vspace{1ex} \noindent{\bf #1.}}

\newcommand{\ucla}{\raisebox{5pt}{\includegraphics[scale=0.038]{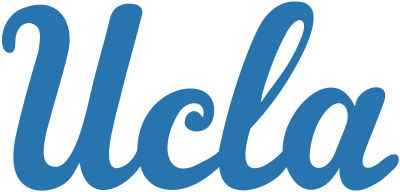}}}
\newcommand{\usc}{\raisebox{5pt}{\includegraphics[scale=0.0125]{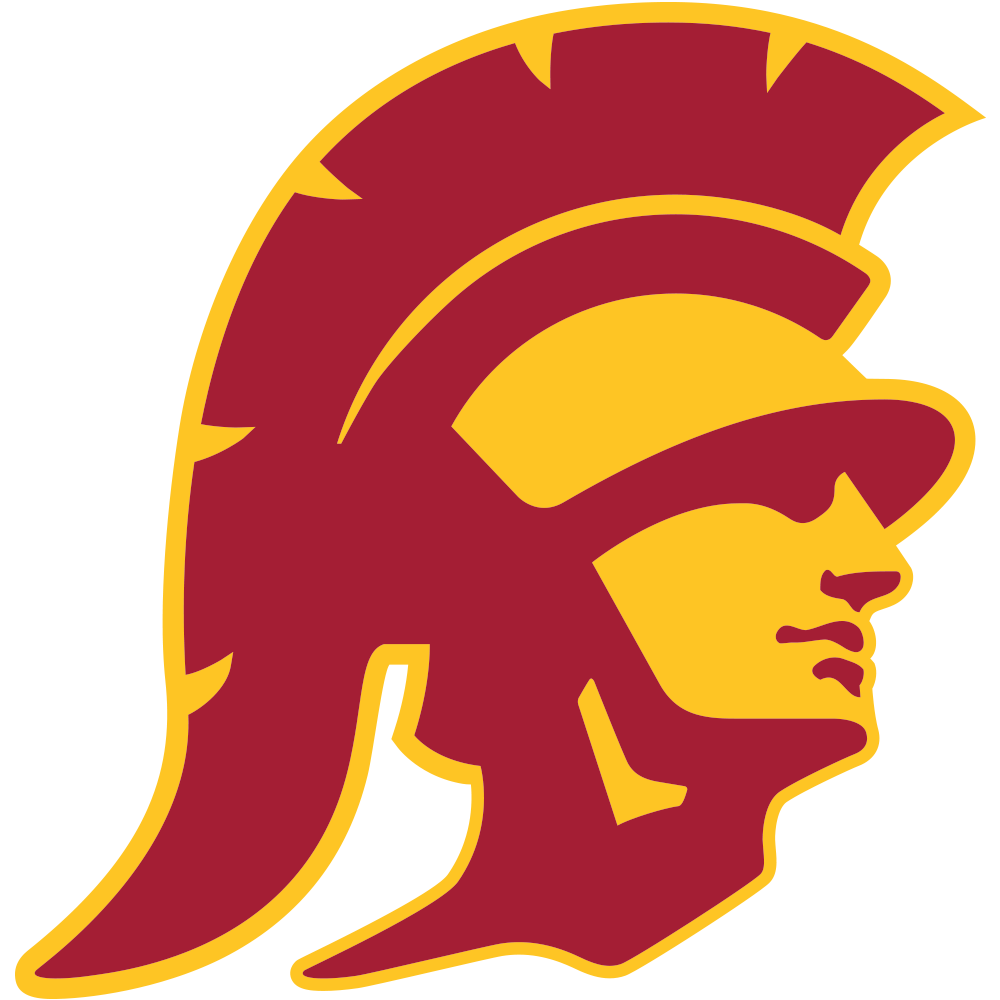}}}
\newcommand{\ucd}{\raisebox{5pt}{\includegraphics[scale=0.006]{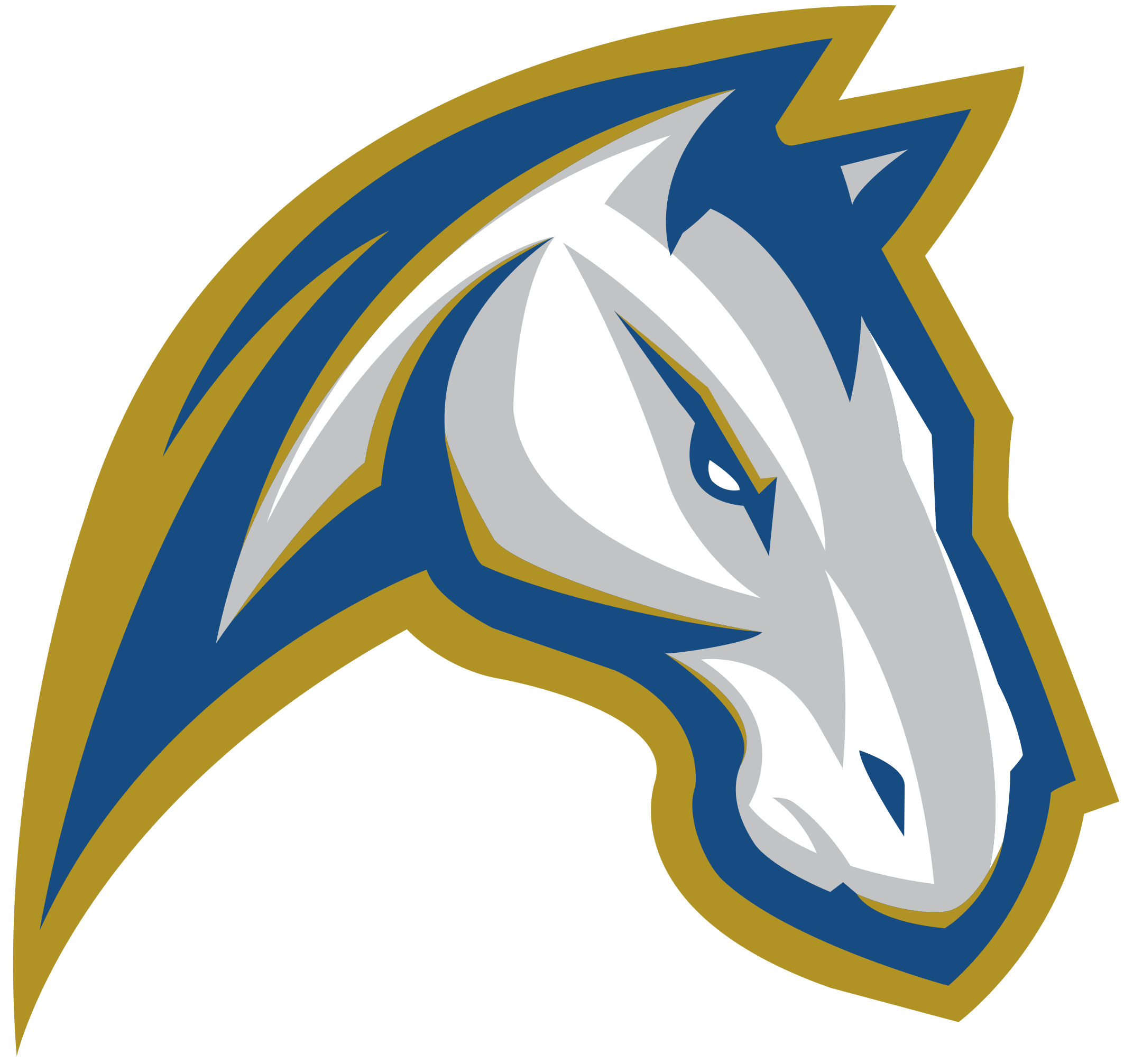}}}

\usepackage{xspace}
\newcommand{\MODEL}{\mbox{\textsc{MonoPara}}\xspace}
\newcommand{\tabincell}[2]{\begin{tabular}{@{}#1@{}}#2\end{tabular}}

\title{Monotonic Paraphrasing \\ Improves Generalization of Language Model Prompting}

\author{
Qin Liu\ucd~~
Fei Wang\usc~~
Nan Xu\usc~~
Tianyi Yan \usc~~
Tao Meng \ucla~~
Muhao Chen\ucd\\
{\ucd}UC Davis;\;{\usc}USC;\;{\ucla}UCLA\\
\texttt{\{qinli, muhchen\}@ucdavis.edu};\\
\texttt{\{fwang598, nanx, tianyiy\}@usc.edu};~~~\texttt{tmeng@cs.ucla.edu}\\
  }

\begin{document}
\maketitle

\begin{abstract}

Performance of large language models (LLMs) may vary with different prompts or instructions for even the same task. One commonly recognized factor for this phenomenon is the model's familiarity with the given prompt or instruction, which is typically estimated by its perplexity.
However, finding the prompt with the lowest perplexity is challenging, given the enormous space of possible prompting phrases. 
In this paper, we propose monotonic paraphrasing (\MODEL), an end-to-end decoding strategy that paraphrases given prompts or instructions into their lower perplexity counterparts based on an ensemble of a paraphrase LM for prompt (or instruction) rewriting, and a target LM (i.e. the prompt or instruction executor) that constrains the generation for lower perplexity.
The ensemble decoding process can efficiently paraphrase the original prompt without altering its semantic meaning, while 
monotonically decreasing the perplexity of each generation as calculated by the target LM. 
We explore in detail both greedy and search-based decoding as two alternative decoding schemes of \MODEL.
Notably,
\MODEL does not require any training and can monotonically lower the perplexity of the paraphrased prompt or instruction, leading to improved 
performance of zero-shot LM prompting as evaluated on a wide selection of tasks.
In addition, \MODEL is also shown to effectively improve LMs' generalization on perturbed and unseen task instructions.\footnote{Our code is available at \url{https://github.com/luka-group/MonoPara}.}

\end{abstract}

\section{Introduction}
Large language models (LLMs) have demonstrated remarkable proficiency in zero-shot decision making~\cite{gonen-etal-2023-demystifying,schick-schutze-2021-just,brown2020language} and instruction following~\cite{jiang2023mistral,kopf2023openassistant,touvron2023llama,taori2023stanford,chiang2023vicuna,ouyang2022training}. However, there can be significant variance in the performance of seemingly similar prompts~\cite{zhao2021calibrate,lu-etal-2022-fantastically, webson-pavlick-2022-prompt, gonen-etal-2023-demystifying,yan-etal-2024-contrastive}. Despite efforts of studies on prompting LMs~\cite{shin-etal-2020-autoprompt,li-liang-2021-prefix,gao-etal-2021-making, ding-etal-2022-openprompt, sanh2021multitask, kojima2022large}, it is still challenging to develop high-quality prompts that can induce better performance for varying tasks on evolving models in an effort-saving manner.

One consensus reached by recent studies is the inverse relationship between a prompt's perplexity and its task performance~\cite{m-etal-2023-ctqscorer,gonen-etal-2023-demystifying}. This stems from the intuition that the frequency of a prompt (or an instruction)\footnote{We use the terms ``prompt'' and ``instruction'' interchangeably in this paper as they both refer to a natural language command for zero-shot inference.} appearing in the (pre-)training data positively influences the model's familiarity with it, thereby enhancing its capability to perform the described task \cite{iter2023context, gonen-etal-2023-demystifying, wang-etal-2023-readprompt, lee2023crafting}.
As a result, existing attempts 
use specific criteria, especially perplexity
for selecting prompts from a collection of candidates.
For example, \citet{gonen-etal-2023-demystifying} builds a large prompt pool for each task and selects the one with the lowest perplexity.
However, the performance of such \emph{rewrite-then-select} methods \cite{jiang-etal-2020-know,zhou2022large,gonen-etal-2023-demystifying,prasad-etal-2023-grips} is limited by the size, quality, and even availability of the candidate pool.
Searching the prompt with the lowest perplexity for a particular task remains challenging due to the vast expanse of potential prompts. 

To address this challenge, we propose a novel progressive \emph{end-to-end} prompt refining approach, namely monotonic paraphrasing (\MODEL), that can \textit{proactively} rewrite the given prompt into its low-perplexity counterpart without compromising its expressiveness of the task. 
During prompt (or instruction) refinement, \MODEL conducts an ensemble decoding process of a paraphrasing model together with the target LM.
The paraphrase model is instructed to rewrite the given prompt of a specific task, while the target model, which is later on used for task inference, provides a constraint that aims to lower the perplexity of the generation. \MODEL iteratively decodes tokens based on the ensemble of these two models. Intuitively, the paraphrase model can genuinely paraphrase the original prompt without altering its semantic meaning and remain instructive for the given task, while the target model restricts the search space of paraphrased tokens to those with low perplexity, resulting in a task prompt that the target model is more familiar with.

Further, for the decoding process of \MODEL, we explore two decoding schemes.
In addition to \emph{greedy decoding} (\Cref{method:ensemble}) 
that iteratively generates the next token based on the weighted 
combination of predicted probabilities
from both models,
we also explored \emph{search-based decoding} (\Cref{method:search}).
The search-based decoding scheme follows the look-ahead decoding paradigm \cite{lu2022neurologic}, which keeps several sequence candidates and maintains the one with the lowest perplexity scored by the target model.
Compared to recent prompt refinement approaches~\cite{gonen-etal-2023-demystifying,m-etal-2023-ctqscorer,shum2023automatic}, \MODEL needs zero training effort and far less computational cost,
and do not require the pre-existence of multiple prompt or instruction candidates \cite{gonen-etal-2023-demystifying}. We find that by an ensemble of two LMs, we are able to consistently decode lower perplexity prompts, which would, in turn, bring about better generalization of LMs on downstream tasks.

Our contributions are three-fold. First, we propose \MODEL, a prompt refinement method that paraphrases prompts to be more familiar with the target model for boosted generalization and task performance. Second, we explore and compare two distinct decoding strategies that proactively refine prompts by monotonically lowering their perplexity and maintaining their expressiveness of the task. Third, we conduct experimentation for both prompt refinement for regular LMs and instruction refinement for instruction-tuned LMs to illustrate that monotonic paraphrasing improves the generalization of LM prompting.










\begin{figure}[t]
    \centering
    \includegraphics[width=0.48\textwidth]{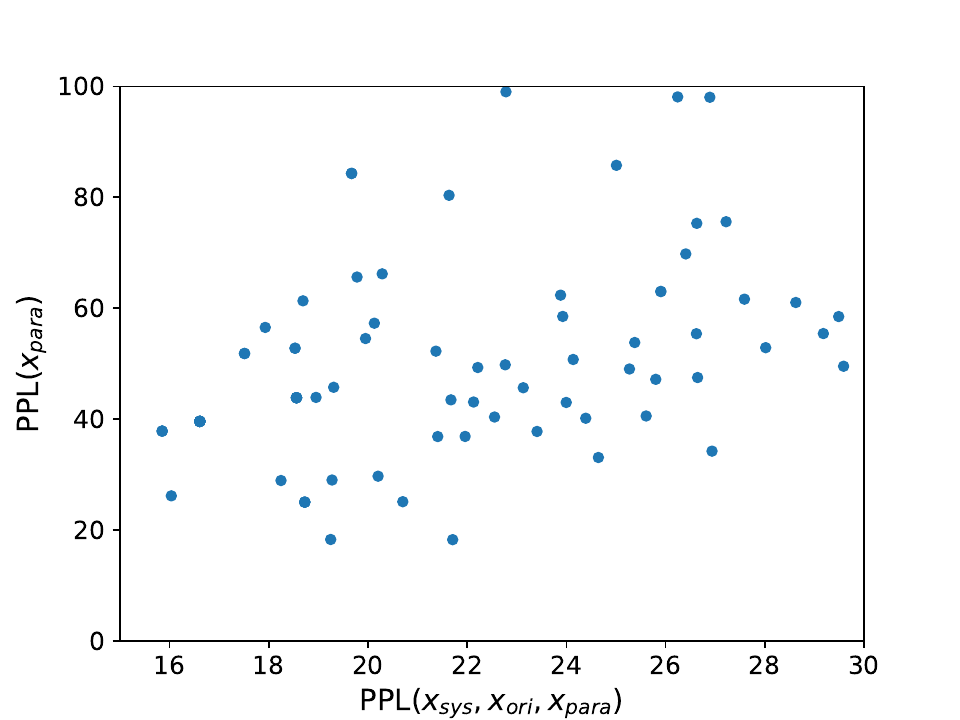}
    \caption{Perplexity of $x_{para}$ as the output paraphrase of $P_{para}$ vs. as the input prompt of $P_{tar}$ for the AG News dataset with Mistral 7B as both $P_{para}$ and $P_{tar}$. Each point stands for a different prompt $x_{para}$. A low-perplexity paraphrase does not necessarily result in a low-perplexity prompt for the target model.}
    \label{fig:scatter}
\end{figure}

\begin{figure*}[t]
    \centering
    \includegraphics[width=0.9\textwidth]{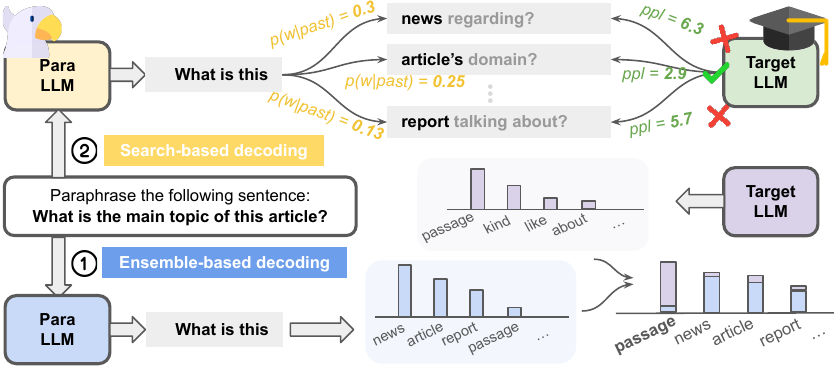}
    \caption{Two explored decoding schemes of \MODEL. Ensemble-based decoding (bottom) combines the token probabilities from the paraphrase model and the target model in each decoding step. Search-based decoding (top) further leverages look-ahead decoding to consider the potential future impact of current choices.}
    \label{fig:method}
\end{figure*}

\section{Monotonic Paraphrasing}
We propose \MODEL as an ensemble-based decoding method that can refine a given prompt by paraphrasing it into a low-perplexity counterpart. We first provide the intuition of \MODEL in \Cref{method:intuition} and the basics of prompt paraphrasing in \Cref{method:paraphrase}. Then, we formally define two decoding schemes of \MODEL in \Cref{method:ensemble} and \Cref{method:search}.

\subsection{Intuition and Methodology}
\label{method:intuition}


Assume that the user wants to address a task specified by a given prompt $x_{ori}$ using a target LM $P_{tar}$.
\MODEL seeks to refine the given prompt to its low-perplexity counterpart $x_{para}$ (i.e. a more \emph{familiar} prompt to $P_{tar}$), thereby enhancing model performance.
However, simply paraphrasing the original prompt with a paraphrase model $P_{para}$ does not necessarily result in a low-perplexity counterpart. 
This discrepancy arises from a mismatch between the perplexity of the output $x_{para}$ from the paraphrase model and the perplexity observed when using $x_{para}$ as the input prompt for the target model.
As shown in \Cref{fig:scatter}, there is a lack of significant correlation between the perplexity of $x_{para}$ as output in $P_{para}$ and that of $x_{para}$ as input in $P_{tar}$, even if we use the same model as the paraphrase and target model ($P_{para} = P_{tar}$).  
This observation indicates that the minimal perplexity of a paraphrased prompt could not be achieved by the paraphrase model alone.


Based on the above fact, the goal of \MODEL is to take on the semantic hints from the paraphrase model while following the constraint of perplexity from the target model.
Specifically, the paraphrase model generates paraphrases for a given prompt, while the target model constrains the paraphrase perplexity during the generation process. With an ensemble of these two models, we can leverage the strengths of both: the guided content generation capability of the paraphrase model and the low-perplexity prompt constraint of the target model. \MODEL is thus able to produce paraphrases with lower perplexity by synergizing the benefits of both models.



\subsection{Prompt Paraphrasing}
\label{method:paraphrase}
We first introduce prompt paraphrasing without perplexity constraint.
We consider using an LLM for prompt paraphrasing in a zero-shot manner, where the model is instructed to paraphrase the original prompt $x_{ori}$ to a fluent and coherent counterpart $x_{para}$ with a \emph{system prompt} $x_{sys}$, such as the following one:
\begin{tcolorbox}[colback=white, colframe=black!75!black, boxrule=0.5pt, sharp corners, title=The System Prompt for Paraphrasing]
\small
Generate ONE paraphrase of the following sentence.
\end{tcolorbox}

\noindent
Specifically, we denote the input of the paraphrase model, including the concatenation of the system prompt $x_{sys}$ and the original prompt $x_{ori}$ that the model is instructed to rewrite, as $[x_{sys}, x_{ori}] = (x_1, \dots, x_n)$, 
where $x_i$ is a token in the vocabulary $V$. Suppose the decoder of a pre-trained autoregressive language model $P_{para}$ generates continuations of length $m$ as the requested paraphrase, denoted as $x_{para} = (x_{n+1}, \dots, x_{n+m})$, based on the system prompt $x_{sys}$ and original prompt $x_{ori}$.
At decoding time, $P_{para}$ iteratively decodes one token at a time by conditioning on the preceding context:

\begin{equation*}
    P_{para}(x_{para} | x_{sys}, x_{ori}) = \prod_{i=n+1}^{n+m} P_{para} (x_i | x_{<i}),
\end{equation*}
where $P_{para}(x_i | x_{<i})$ is the predicted next token probability. 


\subsection{Ensemble-based (Greedy) \MODEL}
\label{method:ensemble}
To introduce perplexity constraint from the target model to the paraphrase model, here we introduce the ensemble-based greedy decoding for monotonic paraphrasing (as shown in the lower half of \Cref{fig:method}).
Based on the intuition delivered in \Cref{method:intuition}, \MODEL decodes the paraphrase iteratively, and the next token is selected based on the combination of predicted probabilities from the two LMs:
\begin{align*}
x_{next} &= \underset{x_i \in V}{\mathrm{argmax}} (\alpha \cdot \log P_{tar}(x_i | x_{gen}) \\ &+(1-\alpha) \cdot\log P_{para}(x_i | x_{sys}, x_{ori}, x_{gen})),
\end{align*}
where 
$x_{gen}$ denotes the sequence of generated tokens, and $\alpha$ determines the coefficient between the two models (\Cref{alg:ensemble}).

By combining the prediction probabilities of both models, the ensemble can navigate the complex landscape of language generation more effectively and precisely. The paraphrase model $P_{para}$ keeps the 
rewritten prompt (or instruction) remains with 
the original intent, while the target model $P_{tar}$ evaluates the paraphrase's likelihood or confidence, up-weighting any generation samples that lead towards low-perplexity outcomes.
Based on the ensemble, the next token is selected by maximizing the weighted sum of logarithmic probabilities from both models, considering both the instruction and the generated text thus far.

As a coefficient, $\alpha$ balances the contribution of each model, ensuring that the final output satisfies both \textbf{semantic fidelity} to the prompt intent and \textbf{linguistic familiarity} to the target model.
Higher $\alpha$ increases the weight of the logarithmic probability derived from the target model $P_{tar}$ and cuts down on the semantic hints for $P_{para}$, which raises the risk of sacrificing the semantic fidelity of the generation.
Conversely, a lower $\alpha$ shifts the emphasis towards $P_{para}$, placing greater value on ensuring that the generated text closely follows the instructional context and semantic intent from $P_{para}$. In this case, the decoding process lacks the constraint of perplexity from $P_{tar}$ and sacrifices the linguistic familiarity of the generated paraphrase for the target model.

\begin{algorithm}[t]
\caption{Ensemble-based Decoding}
\label{alg:ensemble}
\begin{algorithmic}[1]
\Require
\Statex \(x_{sys}\) \Comment{System prompt for paraphrase}
\Statex \(V\) \Comment{Vocabulary}
\Statex \(P_{tar}, P_{para}\) \Comment{Target and paraphrase models}
\Statex \(\alpha\) \Comment{Weighting factor}
\Ensure
\Statex \(x_{gen}\) \Comment{Generated text sequence}

\State Initialize \(x_{gen} \gets \emptyset\)
\State \(x_{next} \gets \underset{x_i \in V}{\mathrm{argmax}}\ P_{para}(x_i | x_{sys} \oplus x_{ori})\) \Comment{Generate first token with \(P_{para}\)}
\State \(x_{gen} \gets x_{gen} \oplus x_{next}\)

\While{not end of sequence}
    \State Compute weighted log probabilities for each \(x_i \in V\):
    \State \(score_{tar} \gets \alpha \cdot \log P_{tar}(x_i | x_{gen})\)
    \State \(score_{para} \gets (1-\alpha) \cdot \log P_{para}(x_i | x_{sys} \oplus x_{ori} \oplus x_{gen})\)
    \State \(x_{next} \gets \underset{x_i \in V}{\mathrm{argmax}}\ (score_{tar} + score_{para})\)
    \State \(x_{gen} \gets x_{gen} \oplus x_{next}\)
\EndWhile

\State \Return \(x_{gen}\)
\end{algorithmic}
\end{algorithm}

\subsection{Search-based \MODEL}
\label{method:search}

To further enhance the efficiency of decoding prompts of lower perplexity, 
we define the search-based decoding strategy for \MODEL (as shown in the upper half of \Cref{fig:method}). To take more steps of token generation into account and expand the search space, we leverage look-ahead decoding to search for the sequence that exhibits the lowest perplexity from an even broader candidate space. Following \citet{lu-etal-2022-neurologic}, each step of our search-based decoding consists of (\romannumeral 1) expanding a set of candidate next-tokens, (\romannumeral 2) scoring each candidate, and (\romannumeral 3) selecting the $k$ best candidates (\Cref{alg:search} in \Cref{append:alg2}):
\begin{align*}
    X'_m &= \{\mathrm{x}_{<m} \circ x_m | \mathrm{x}_{<m} \in X_{m-1}, x_m \in V\}, \\
    X_m &= \underset{(\mathrm{x}_{<m},x_m) \in X'_m}{\mathrm{\arg\,topk}} \{f(\mathrm{x}_{<m},x_m)\},
\end{align*}
where $x_m$ is a candidate predicted by paraphrase model, and $f(\cdot)$ is a scoring function that returns the perplexity of $\mathrm{x}_{<m} \circ x_m$ evaluated by the target model:
\begin{align}
    f(\mathrm{x}_{<m},x_m) = \left( \prod_{i=1}^{m} \frac{1}{P_{tar}(x_i | x_{<i})} \right)^{\frac{1}{m}}.
\label{equ:ppl}
\end{align}

\noindent
Specifically, at each step $m$ of the decoding process, the current set of sequences $X_{m-1}$ is expanded by appending the most probable next-token $x_m$, predicted by the paraphrase model $P_{para}$, from vocabulary $V$ to each sequence $\mathrm{x}_{<m}$. This results in a temporary expanded set of candidate sequences $X'_m$.
Each candidate sequence in $X'_m$ is then evaluated by the scoring function $f(\cdot)$ shown in \Cref{equ:ppl}, which calculates the perplexity of a sequence by the target model $P_{tar}$.
Since a paraphrase of lower perplexity is preferred,  the $k$ sequences with the best scores (i.e., lowest perplexity) are selected to form the set $X_m$, narrowing down the choices to the $k$ most promising sequences for continuation.

In comparison to greedy decoding,
this look-ahead decoding mechanism evaluates a broader range of potential future impact from current choices (i.e., the selection of $x_m$), thereby allowing for a more informed and potentially more accurate selection of candidates. 

\section{Experiment}
\label{sec:exp_setting}
In this section, we demonstrate two distinct experimental settings for evaluation.
In \Cref{exp:prompt}, we examine \MODEL's effectiveness on prompt refinement for eliciting better task performance.
In \Cref{exp:robustness}, we evaluate \MODEL's effect on enhancing model robustness and generalization under instruction perturbations.

\subsection{Task \uppercase\expandafter{\romannumeral 1}: Prompt Refinement}
\label{exp:prompt}


\paragraph{Task Description}
To inspect if monotonic paraphrasing can refine a prompt and elicit better performance on prompting an LLM as designed, we use \textsc{Super-NaturalInstruction} (\textsc{Sup-NatInst} for short, \citealp{wang-etal-2022-super}).
This benchmark consists of 1,616 diverse NLP tasks and their expert-written prompts for evaluating the zero-shot generalizability of LLMs on a variety of NLP tasks. In \textsc{Sup-NatInst}, each task is paired up with an instruction that consists of the task definition for mapping an input text to a task output and several examples for demonstrating the desired or undesired output. To make our evaluation more challenging, we only use the task definition as a prompt for the target model to perform the task. During test time, the target model is prompted with a concatenation of \textsc{Sup-NatInst} task description (or its paraphrase), test sample, and multiple choices of answers. Since \textsc{Sup-NatInst} provides only one description of definition for each task, we use GPT4 \cite{achiam2023gpt} to generate $4$ more \textsc{Sup-NatInst}-like task descriptions for each involved task, which results in $5$ prompts for each task in total.

\paragraph{Datasets}
Following \citet{gonen-etal-2023-demystifying}, we choose $7$ classification tasks from Huggingface Datasets \footnote{https://huggingface.co/docs/datasets/index} to have a set of diverse tasks: (\romannumeral 1) GLUE Cola \cite{warstadt-etal-2019-neural} for grammatical acceptability discrimination; (\romannumeral 2) Newspop \cite{moniz2018multi} for news classification; (\romannumeral 3) AG News \cite{NIPS2015_250cf8b5} for news classification; (\romannumeral 4) IMDB \cite{maas-etal-2011-learning} for movie review sentiment analysis; (\romannumeral 5) DBpedia \cite{lehmann2015dbpedia} for topic classification; (\romannumeral 6) Emotion \cite{saravia-etal-2018-carer} for tweet emotion classification; and (\romannumeral 7) Tweet Offensive \cite{barbieri-etal-2020-tweeteval} for offensive tweet discrimination. We sample 1,000 examples from each dataset for prompt evaluation.
For these tasks, the prompt follows the input, and at the end of each prompt, we add the choices of classes following \citet{gonen-etal-2023-demystifying}: ``Choices: X, Y, Z. Answer: '' as it helps with accuracy.

\paragraph{Evaluation Metrics}
To inspect whether monotonic paraphrasing can refine a prompt into its low-perplexity counterpart and further enhance the generalization of LLM prompting, we use PPL (perplexity), Acc (accuracy), and BS (BERTScore, \citealp{Zhang*2020BERTScore:}) for evaluation.
(1) Perplexity.
Following \citet{gonen-etal-2023-demystifying}, we define the perplexity of the prompt as the perplexity of the full prompt sequence, including the input itself and the choices of labels.
To avoid noise when computing perplexity, the prompts are instantiated with 1,000 random examples of the dataset (we keep the same selected examples for performance evaluation), and the perplexity is averaged across all instantiated prompts.
The correlation between the perplexity of a standalone prompt and the average perplexity over instantiated ones is illustrated in \Cref{sec:gap}.
(2) BERTScore.
As a paraphrasing scheme, \MODEL is also evaluated by its performance in semantic alignment with the source sentence based on pre-trained BERT \cite{devlin-etal-2019-bert}.
(3) Accuracy.
To compute accuracy, for each test sample, we obtain the model's predicted probability of all classes and choose the highest ranking class as the prediction of the model. The accuracy is calculated over the 1,000 samples for each task.

\begin{table*}[t]
\setlength\tabcolsep{2pt}
\centering
\small
\begin{tabular}{llccccccc}
\toprule
\multicolumn{1}{l|}{\textbf{Methods}} & \multicolumn{1}{l|}{\textbf{Metric}} & \textbf{GLUE Cola} & \textbf{Newspop} & \textbf{AG News} & \textbf{IMDB} & \textbf{DBpedia} & \textbf{Emotion} & \textbf{Tweet Offensive} \\ \midrule
\multicolumn{9}{c}{\textbf{Mistral 7B}} \\ \midrule
\multicolumn{1}{l|}{} & \multicolumn{1}{l|}{\cellcolor[HTML]{EFEFEF}Acc} & \cellcolor[HTML]{EFEFEF}61.76 ± 2.93 & \cellcolor[HTML]{EFEFEF}85.28 ± 3.45 & \cellcolor[HTML]{EFEFEF}67.68 ± 3.47 & \cellcolor[HTML]{EFEFEF}84.12 ± 4.89 & \cellcolor[HTML]{EFEFEF}70.82 ± 2.27 & \cellcolor[HTML]{EFEFEF}44.32 ± 1.57 & \cellcolor[HTML]{EFEFEF}\textbf{67.04} ± 0.95 \\
\multicolumn{1}{l|}{\multirow{-2}{*}{Ori.}} & \multicolumn{1}{l|}{PPL} & 15.72 ± 4.60 & 12.60 ± 2.40 & 12.74 ± 1.07 & 16.41 ± 0.80 & \textcolor{blue}{8.10} ± 0.22 & 16.48 ± 2.14 & 19.25 ± 3.43 \\ \midrule
\multicolumn{1}{l|}{} & \multicolumn{1}{l|}{\cellcolor[HTML]{EFEFEF}Acc} & \cellcolor[HTML]{EFEFEF}43.42 ± 1.16 & \cellcolor[HTML]{EFEFEF}81.08 ± 7.2 & \cellcolor[HTML]{EFEFEF}\textbf{73.88} ± 3.18 & \cellcolor[HTML]{EFEFEF}86.00 ± 0.81 & \cellcolor[HTML]{EFEFEF}71.02 ± 1.70 & \cellcolor[HTML]{EFEFEF}46.80 ± 1.12 & \cellcolor[HTML]{EFEFEF}65.06 ± 1.04 \\
\multicolumn{1}{l|}{\multirow{-2}{*}{SPELL}} & \multicolumn{1}{l|}{PPL} & \textcolor{blue}{8.24} ± 0.88 & 15.05 ± 4.15 & 15.99 ± 3.77 & 15.76 ± 1.42 & 25.36 ± 6.80 & 15.67 ± 5.21 & 14.31 ± 0.93 \\ \midrule
\multicolumn{1}{l|}{} & \multicolumn{1}{l|}{\cellcolor[HTML]{EFEFEF}Acc} & \cellcolor[HTML]{EFEFEF}39.48 ± 8.19 & \cellcolor[HTML]{EFEFEF}76.72 ± 16.54 & \cellcolor[HTML]{EFEFEF}71.60 ± 2.70 & \cellcolor[HTML]{EFEFEF}86.60 ± 4.86 & \cellcolor[HTML]{EFEFEF}70.22 ± 5.71 & \cellcolor[HTML]{EFEFEF}43.65 ± 3.41 & \cellcolor[HTML]{EFEFEF}65.04 ± 1.46 \\
\multicolumn{1}{l|}{} & \multicolumn{1}{l|}{PPL} & 78.01 ± 11.22 & 49.37 ± 7.81 & 28.37 ± 2.58 & 22.27 ± 0.98 & 28.34 ± 1.36 & 46.53 ± 5.30 & 53.65 ± 1.92 \\
\multicolumn{1}{l|}{\multirow{-3}{*}{Para}} & \multicolumn{1}{l|}{BS} & 88.88 ± 0.74 & 87.47 ± 0.84 & 86.32 ± 2.08 & \textcolor{orange}{90.23} ± 0.99 & 86.96 ± 0.93 & 90.08 ± 0.35 & 89.93 ± 0.67 \\ \midrule
\multicolumn{1}{l|}{} & \multicolumn{1}{l|}{\cellcolor[HTML]{EFEFEF}Acc} & \cellcolor[HTML]{EFEFEF}\textbf{63.96} ± 3.88 & \cellcolor[HTML]{EFEFEF}\textbf{88.96} ± 3.04 & \cellcolor[HTML]{EFEFEF}72.20 ± 3.87 & \cellcolor[HTML]{EFEFEF}\textbf{87.58} ± 3.27 & \cellcolor[HTML]{EFEFEF}72.07 ± 3.54 & \cellcolor[HTML]{EFEFEF}\textbf{47.76} ± 3.72 & \cellcolor[HTML]{EFEFEF}67.00 ± 1.21 \\
\multicolumn{1}{l|}{} & \multicolumn{1}{l|}{PPL} & 14.89 ± 2.75 & \textcolor{blue}{8.33} ± 1.30 & 10.07 ± 2.93 & \textcolor{blue}{13.05} ± 1.86 & 8.85 ± 0.88 & \textcolor{blue}{13.67} ± 1.87 & \textcolor{blue}{14.11} ± 3.53 \\
\multicolumn{1}{l|}{\multirow{-3}{*}{\textbf{Mono-E}}} & \multicolumn{1}{l|}{BS} & \textcolor{orange}{93.03} ± 2.15 & 92.85 ± 2.35 & 88.70 ± 2.35 & 89.35 ± 2.60 & 93.21 ± 1.16 & 92.74 ± 2.75 & \textcolor{orange}{91.73} ± 1.32 \\ \midrule
\multicolumn{1}{l|}{} & \multicolumn{1}{l|}{\cellcolor[HTML]{EFEFEF}Acc} & \cellcolor[HTML]{EFEFEF}62.31 ± 6.17 & \cellcolor[HTML]{EFEFEF}86.34 ± 5.16 & \cellcolor[HTML]{EFEFEF}70.80 ± 3.44 & \cellcolor[HTML]{EFEFEF}83.48 ± 2.87 & \cellcolor[HTML]{EFEFEF}\textbf{72.46} ± 3.00 & \cellcolor[HTML]{EFEFEF}44.26 ± 3.05 & \cellcolor[HTML]{EFEFEF}65.66 ± 2.43 \\
\multicolumn{1}{l|}{} & \multicolumn{1}{l|}{PPL} & 15.85 ± 4.88 & 11.95 ± 2.09 & \textcolor{blue}{9.89} ± 1.31 & 16.38 ± 1.63 & 8.63 ± 0.41 & 15.03 ± 1.65 & 18.35 ± 4.07 \\
\multicolumn{1}{l|}{\multirow{-3}{*}{\textbf{Mono-S}}} & \multicolumn{1}{l|}{BS} & 92.47 ± 2.00 & \textcolor{orange}{93.78} ± 2.68 & \textcolor{orange}{91.57} ± 2.23 & 89.66 ± 3.25 & \textcolor{orange}{95.87} ± 1.20 & \textcolor{orange}{93.42} ± 1.02 & 91.62 ± 2.48 \\ \midrule
\multicolumn{9}{c}{\textbf{Starling 7B}} \\ \midrule
\multicolumn{1}{l|}{} & \multicolumn{1}{l|}{\cellcolor[HTML]{EFEFEF}Acc} &
\cellcolor[HTML]{EFEFEF}55.18 ± 12.43 & \cellcolor[HTML]{EFEFEF}93.78 ± 1.18 & \cellcolor[HTML]{EFEFEF}\textbf{88.44} ± 1.18 & \cellcolor[HTML]{EFEFEF}96.46 ± 0.17 & \cellcolor[HTML]{EFEFEF}\textbf{94.06} ± 0.37 & \cellcolor[HTML]{EFEFEF}55.48 ± 0.91 & \cellcolor[HTML]{EFEFEF}64.62 ± 0.18 \\
\multicolumn{1}{l|}{\multirow{-2}{*}{Ori.}} & \multicolumn{1}{l|}{PPL} & 16.38 ± 4.18 & 12.77 ± 2.59 & 12.10 ± 0.91 & 15.56 ± 0.57 & \textcolor{blue}{8.11} ± 0.22 & 17.90 ± 2.82 & 20.41 ± 3.90 \\ \midrule
\multicolumn{1}{l|}{} & \multicolumn{1}{l|}{\cellcolor[HTML]{EFEFEF}Acc} & \cellcolor[HTML]{EFEFEF}46.46 ± 6.70 & \cellcolor[HTML]{EFEFEF}89.84 ± 0.91 & \cellcolor[HTML]{EFEFEF}83.72 ± 0.26 & \cellcolor[HTML]{EFEFEF}92.84 ± 2.7 & \cellcolor[HTML]{EFEFEF}92.02 ± 0.64 & \cellcolor[HTML]{EFEFEF}54.82 ± 1.81 & \cellcolor[HTML]{EFEFEF}64.20 ± 0.17 \\
\multicolumn{1}{l|}{\multirow{-2}{*}{SPELL}} & \multicolumn{1}{l|}{PPL} & 44.96 ± 18.23 & 54.60 ± 18.72 & 16.81 ± 0.91 & 56.34 ± 15.67 & 28.66 ± 9.47 & 32.79 ± 8.53 & 33.56 ± 3.83 \\ \midrule
\multicolumn{1}{l|}{} & \multicolumn{1}{l|}{\cellcolor[HTML]{EFEFEF}Acc} & \cellcolor[HTML]{EFEFEF}40.38 ± 15.49 & \cellcolor[HTML]{EFEFEF}82.67 ± 19.63 & \cellcolor[HTML]{EFEFEF}86.39 ± 2.92 & \cellcolor[HTML]{EFEFEF}\textbf{96.66} ± 0.34 & \cellcolor[HTML]{EFEFEF}93.10 ± 1.79 & \cellcolor[HTML]{EFEFEF}51.10 ± 10.18 & \cellcolor[HTML]{EFEFEF}64.44 ± 0.08 \\
\multicolumn{1}{l|}{} & \multicolumn{1}{l|}{PPL} & 70.50 ± 16.38 & 54.41 ± 8.80 & 30.67 ± 2.81 & 21.78 ± 1.37 & 26.56 ± 1.39 & 58.87 ± 4.44 & \textcolor{blue}{11.71} ± 0.86 \\
\multicolumn{1}{l|}{\multirow{-3}{*}{Para}} & \multicolumn{1}{l|}{BS} & 90.35 ± 0.67 & 88.35 ± 0.51 & 88.46 ± 0.98 & 90.04 ± 1.04 & 87.06 ± 0.79 & 90.38 ± 0.45 & 79.93 ± 11.52 \\ \midrule
\multicolumn{1}{l|}{} & \multicolumn{1}{l|}{\cellcolor[HTML]{EFEFEF}Acc} & \cellcolor[HTML]{EFEFEF}\textbf{69.62} ± 4.12 & \cellcolor[HTML]{EFEFEF}\textbf{95.15} ± 0.59 & \cellcolor[HTML]{EFEFEF}85.40 ± 4.22 & \cellcolor[HTML]{EFEFEF}96.22 ± 0.47 & \cellcolor[HTML]{EFEFEF}94.05 ± 0.42 & \cellcolor[HTML]{EFEFEF}\textbf{56.10} ± 1.08 & \cellcolor[HTML]{EFEFEF}64.45 ± 0.09 \\
\multicolumn{1}{l|}{} & \multicolumn{1}{l|}{PPL} & \textcolor{blue}{8.56} ± 1.00 & \textcolor{blue}{10.12} ± 1.02 & \textcolor{blue}{8.52} ± 0.76 & \textcolor{blue}{12.60} ± 1.57 & 9.58 ± 0.86 & \textcolor{blue}{9.23} ± 0.97 & \textcolor{blue}{11.71} ± 0.86 \\
\multicolumn{1}{l|}{\multirow{-3}{*}{\textbf{Mono-E}}} & \multicolumn{1}{l|}{BS} & \textcolor{orange}{93.00} ± 0.85 & 94.19 ± 0.94 & 91.06 ± 1.51 & 92.85 ± 1.39 & 93.23 ± 0.50 & 93.28 ± 0.26 & 93.69 ± 1.52 \\ \midrule
\multicolumn{1}{l|}{} & \multicolumn{1}{l|}{\cellcolor[HTML]{EFEFEF}Acc} & \cellcolor[HTML]{EFEFEF}50.96 ± 11.52 & \cellcolor[HTML]{EFEFEF}88.20 ± 3.74 & \cellcolor[HTML]{EFEFEF}83.61± 6.30 & \cellcolor[HTML]{EFEFEF}92.96 ± 4.78 & \cellcolor[HTML]{EFEFEF}92.82 ± 2.17 & \cellcolor[HTML]{EFEFEF}54.90 ± 2.32 & \cellcolor[HTML]{EFEFEF}\textbf{66.80} ± 1.22 \\
\multicolumn{1}{l|}{} & \multicolumn{1}{l|}{PPL} & 14.05 ± 3.76 & 12.55 ± 2.07 & 10.61 ± 1.06 & 16.44 ± 0.92 & 8.39 ± 0.53 & 16.28 ± 2.18 & 20.17 ± 3.17 \\
\multicolumn{1}{l|}{\multirow{-3}{*}{\textbf{Mono-S}}} & \multicolumn{1}{l|}{BS} & 92.69 ± 0.99 & \textcolor{orange}{95.83} ± 1.66 & \textcolor{orange}{93.92} ± 2.23 & \textcolor{orange}{93.01} ± 1.25 & \textcolor{orange}{96.99} ± 0.70 & \textcolor{orange}{94.78} ± 1.25 & \textcolor{orange}{94.45} ± 1.43 \\ 
\bottomrule
\end{tabular}
\caption{Results of Task \uppercase\expandafter{\romannumeral 1} (Prompt Refinement). Ori. refers to the original prompts. Para refers to paraphrasing the original prompts without perplexity constraint. Mono-E refers to ensemble-based \MODEL. Mono-S refers to search-based \MODEL. The best accuracy for each task is in \textbf{bold}, while the lowest perplexity and the highest BERTScore for each task is in \textcolor{blue}{blue} and \textcolor{orange}{orange} respectively. The coefficient $\alpha$ is fixed as $0.5$ for all the results.}
\label{tab:main} 
\end{table*}

\paragraph{Models}
We study two auto-regressive models, Mistral-7B\footnote{\url{https://huggingface.co/mistralai/Mistral-7B-Instruct-v0.1}} \cite{jiang2023mistral}and Starling-7B\footnote{\url{https://huggingface.co/berkeley-nest/Starling-LM-7B-alpha}} \cite{starling2023}, for their significant ability in instruction following and paraphrasing. In this paper, we keep the paraphrase model \textbf{the same} as the target model. Please refer to \Cref{append:llama} for the results on Llama-3-8B\footnote{\url{https://huggingface.co/meta-llama/Meta-Llama-3-8B}}~\cite{llama3modelcard}.

\paragraph{Baselines}
We compare our method with two baselines: SPELL \cite{gonen-etal-2023-demystifying} and vanilla LLM-based paraphrasing (Para).
Based on similar intuition that low perplexity and better performance exhibit strong correlation, SPELL (\textbf{S}electing \textbf{P}rompts by \textbf{E}stimating \textbf{L}M \textbf{L}ikelihood) ranks and selects the prompts with the lowest perplexity for a given task after creating a set of prompt candidates manually and expanding them to hundred-scale using automatic paraphrasing and back-translation. We use the prompt candidates provided by \citet{gonen-etal-2023-demystifying} and rank them by the perplexity w.r.t. the target model.
Para paraphrases source prompts with greedy decoding (top-$1$ sampling) by directly prompting the target LLM with the instruction introduced in \Cref{method:paraphrase}.
Besides, we also implement reranking based on the prompt candidates generated by \MODEL for a fair comparison with SPELL. The implementation details and results are discussed in \Cref{sec:reranking}.

\paragraph{Results}
As shown in \Cref{tab:main}, the two variants of \MODEL successfully refine prompts to those with perplexities lower than the original ones, thereby enhancing model performance on 10 out of 14 scenarios (2 models, each on 7 datasets). Specifically, Mono-E (ensembled-based \MODEL) improves the accuracy of Mistral-7B by an average of 2.64\% and that of Starling-7B by an average of 1.85\% on seven different datasets. 
In contrast, directly paraphrasing the prompts without perplexity constraint (Para) results in prompts with higher perplexities than the original ones in 13 out of 14 scenarios, most of which lead to worse task performance. This highlights the importance of incorporating perplexity constraints when paraphrasing prompts. The correlation between task accuracy and prompt perplexity is analyzed in \Cref{append:correlation}.

Besides, our method outperforms SPELL with consistently lower perplexity and superior task performance with the refined prompt. This superiority stems from the fact that, although SPELL explores a vast search space with hundred-scale prompt candidates, the candidates are searched in an ad hoc manner without an optimization goal, leading to inefficiencies in prompt rewriting or paraphrasing. In contrast, our method incorporates the perplexity constraint that leads to optimized target-model perplexity during one single refined prompt generation process, leading to both enhanced task performance and refinement efficiency.

Moreover, \MODEL achieves superior performance in terms of both average prompt perplexity and BERTScore, indicating the synergistic role played by both the target model and the paraphrase model in prompt refinement. During decoding, the target model prioritizes achieving low perplexity, while the paraphrase model preserves the original prompt intent, resulting in a high BERTScore.
We provide a detailed case study on BERTScore of Para and \MODEL in \Cref{append:bertscore} to explain why \MODEL outperforms Para on BERTScore.
Specifically, Mono-E outperforms Mono-S in some cases since Mono-S involves the additional hyperparameter of beam size which we did not tune. Additionally, having additional token options at each decoding step may introduce more randomness, potentially affecting the overall performance. That being said, Mono-S mostly exhibits a higher BERTScore than the simpler method Mono-E, indicating better quality of paraphrasing.

\subsection{Task \uppercase\expandafter{\romannumeral 2}: Robustness Evaluation on Instruction Perturbation}
\label{exp:robustness}



We further evaluate \MODEL's effectiveness in enhancing model robustness against various instruction perturbations at character, word, sentence, and semantic levels using PromptBench \cite{zhu2023promptbench}.

\paragraph{Task Description}
PromptBench introduces perturbation to task instructions of a diverse set of tasks, such as sentiment analysis, grammar correctness, duplicate sentence detection, and natural language inference.
It includes character-level perturbations using DeepWordBug \cite{gao_black-box_2018} to introduce typos, word-level using TextFooler \cite{jin_is_2020} to replace words with contextually similar words, and semantic-level by using prompts following the linguistic behavior of different languages. For semantic-level perturbations, the adversary constructs prompts using various languages, such as Chinese, French, Arabic, Spanish, Japanese, and Korean, and then translates these prompts into English. By exploiting the nuances and idiosyncrasies of different languages during translation, it can introduce subtle ambiguities, grammatical errors, or inconsistencies in the input prompt. The perturbed prompt itself is still in English.


\begin{figure}[t]
    \centering
    \includegraphics[width=0.48\textwidth]{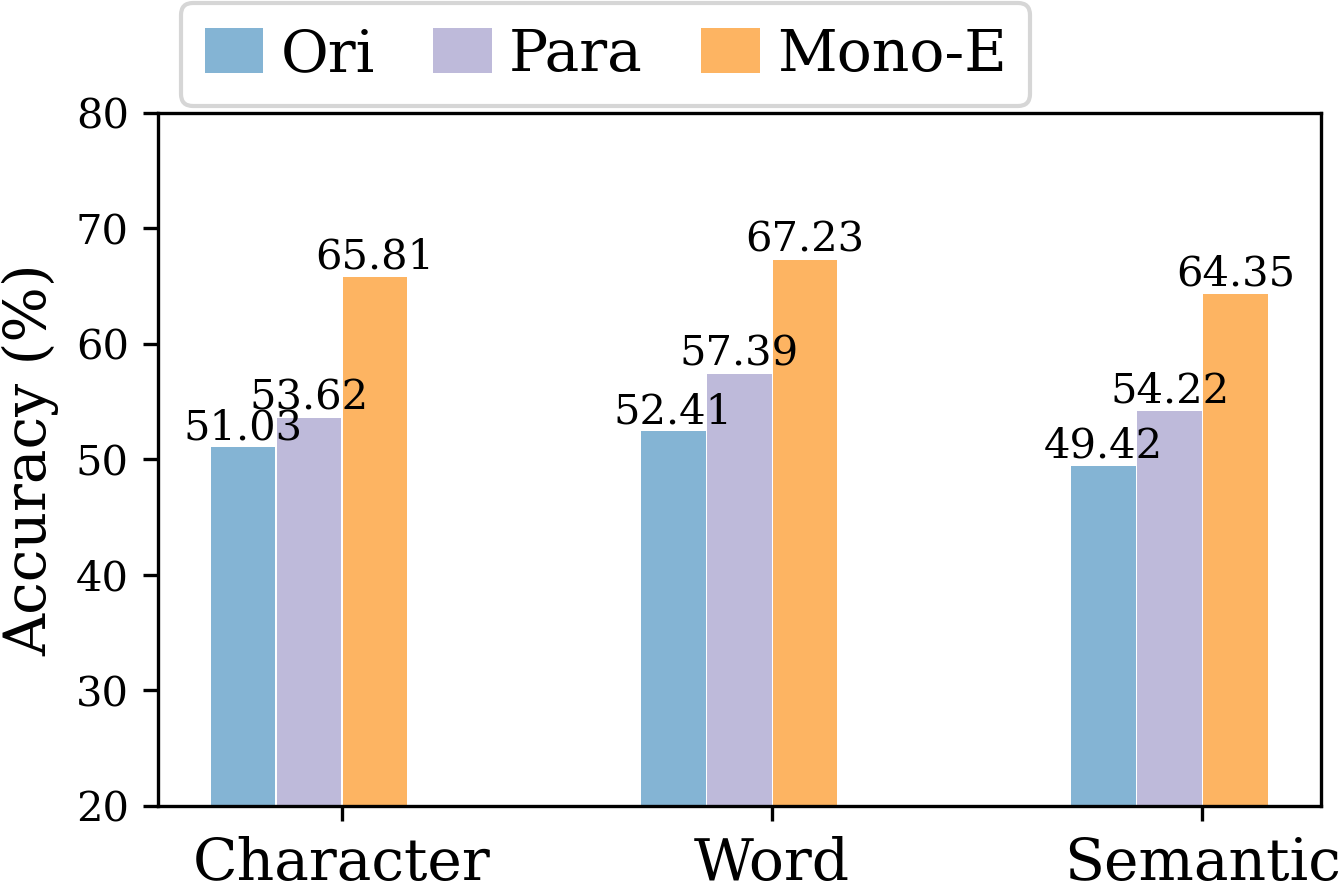}
    \caption{Model's average accuracy across $4$ GLUE datasets, with each dataset having six instructions with perturbation added at character, word, and semantic levels. Mono-E has consistent improvement in accuracy across all types of perturbation compared to vanilla paraphrasing.}
    \label{fig:task2}
\end{figure}

\paragraph{Datasets}
We select $4$ GLUE tasks \cite{wang-etal-2018-glue} from Huggingface Datasets for evaluation: (\romannumeral 1) Cola, which is also used for Task \uppercase\expandafter{\romannumeral 1}; (\romannumeral 2) SST-2 \cite{socher-etal-2013-recursive} for two-way sentiment classification; 
(\romannumeral 3) MNLI \cite{wang-etal-2018-glue} for multi-genre inference identification; and
(\romannumeral 4) MRPC \cite{dolan2005automatically} for paraphrase identification.
We adopt the same instruction format as that of Task \uppercase\expandafter{\romannumeral 1}.

\paragraph{Model}
For this task, we use Alpaca\footnote{\url{https://huggingface.co/tloen/alpaca-lora-7b}. We use a different language model for this task to demonstrate the robustness and versatility of our proposed method across various models.} \cite{taori2023stanford}, a model instruction-tuned from the LLaMA model \cite{touvron2023llama1} on the 52k self-instruct dataset. The paraphrase model is also the same as the target model.

\paragraph{Results}
As shown in \Cref{fig:task2}, we compare the performance of original prompts, prompts paraphrased without perplexity constraint, and our method Mono-E. Across all kinds of instruction perturbations, our method achieves consistent accuracy improvement by up to 10.13\%. This notable increase in performance demonstrates that \MODEL not only refines prompts for improved task performance but also enhances model robustness against various instruction perturbations. We also show the model performance with unperturbed prompts and compare the degradation from perturbation in \Cref{append:taskII} to show the robustness of Mono-E.





\begin{table*}[t]
\setlength\tabcolsep{3pt}
\renewcommand\arraystretch{1.2}
\centering
\small
\begin{tabular}{ll|c|c|c}
\toprule
 &  & Content & BertScore & PPL \\ \midrule
\multicolumn{2}{l|}{Ori Instruction} & \tabincell{c}{Read the provided excerpt and choose between \\ `positive' and `negative' to describe its sentiment Iyk1lJt8yw:} & - & 78.12 \\ \midrule
\multicolumn{2}{l|}{Para} & Determine the sentiment of the given excerpt as either positive or negative & 81.26 & 44.29 \\ \midrule
\multicolumn{1}{l|}{\multirow{3}{*}{Mono}} & $\alpha = 0.2$ & Determine the sentiment of the given text by selecting either `positive' or `negative'. & 86.18 & 24.90 \\ \cline{2-5} 
\multicolumn{1}{l|}{} & $\alpha = 0.5$ & Determine the sentiment of the given text by classifying it as positive or negative. & 80.22 & 16.27 \\ \cline{2-5} 
\multicolumn{1}{l|}{} & $\alpha = 0.7$ & Determine the remainder when 101 is divided by 10.\textbackslash{}n A: 1 & 50.22 & 4.53 \\ \bottomrule
\end{tabular}
\caption{Effect of coefficient $\alpha$ on Mono-E with Mistral-7B model.}
\label{tab:coefficient}
\end{table*}

\subsection{Effect of Coefficient}
\label{analysis:alpha}

We further analyze the effect of coefficient $\alpha$ between the target model and the paraphrase model in \MODEL. \Cref{tab:coefficient} presents a case study of $\alpha$ on the ensemble-based decoding variant (Mono-E) with a prompt under character-level perturbation. Obviously, a larger value of $\alpha$ amplifies the influence of the target model, resulting in a stronger perplexity constraint and monotonically lower perplexity. Consequently, the impact of the paraphrase model is diminished, leading to a lower BERTScore. By adjusting $\alpha$, the user can find a balanced point to achieve better prompts that strike a balance between semantic fidelity to the original prompt intent and linguistic familiarity to the target model. Detailed results and analysis are presented in \Cref{sec:alpha} and \Cref{tab:alpha}.

\section{Related Work}



\stitle{Paraphrasing}
Paraphrase generation has been a longstanding task in NLP. 
Since the era of deep learning, studies have adopted the universal paradigm of training  sequence-to-sequence generation models based on abundant learning resources~\cite{nema-etal-2017-diversity, gupta2018deep, li-etal-2018-paraphrase}.
While more recent works have leveraged more robust pre-trained language models that captured rich supervision signals from various sequence-to-sequence generation tasks \cite{raffel2020exploring, lewis-etal-2020-bart} and achieving even more robust performance of paraphrasing with relatively limited end-task supervision \cite{sun-etal-2021-aesop, meng-etal-2021-conrpg, bui2021generative}.
In addition to these advancements, 
a series of works have added auxiliary supervision signals such as syntactical guidance, lexical regularization, and cross-lingual supervision, to further improve the performance and controllability of paraphrasing in aspects such as syntactic structures, stylistic specification, or textual simplicity
\cite{iyyer-etal-2018-adversarial, li-etal-2019-decomposable, chen-etal-2019-controllable, kumar-etal-2020-syntax, goyal-durrett-2020-neural, hosking-lapata-2021-factorising,chowdhury2022novelty}. 

Different from existing paraphrasing techniques, this work conducts constrained decoding methods that seek to monotonically decrease the perplexity in the process of paraphrasing.
This innovative approach serves as a general approach to rewrite and refine prompts or task instructions towards their more familiar counterparts for LMs, seeking to enhance the generalization of LM prompting without the need for any training effort.

\stitle{Prompt Refinement}
Prompt refinement has been actively explored in recent years
with the aim of selecting, retrieving, or even generating prompts that lead to improved zero-shot performance of LM prompting.
When an end-task objective is known, prior works on prompt refinement approaches often set the end-task performance as the objective, and leverage approaches such as gradient-based search, similar to continuous prompts but with projections onto a discrete vocabulary \cite{shin-etal-2020-autoprompt}. Other works have explored edit-based enumeration \cite{prasad-etal-2023-grips}, reinforcement learning \cite{deng-etal-2022-rlprompt}, and large language model continuation and filtering \cite{zhou2022large}.
As for general-purpose LLMs, a dedicated prompt refinement approach may not assume the pre-existence of any specific end tasks.
Hence, more recent works
\citet{iter2023context, gonen-etal-2023-demystifying, wang-etal-2023-readprompt, lee2023crafting} explored the heuristic criterion based on the familiarity of the language model with the language contained in the prompt, as measured by its perplexity. Lower perplexity prompts are preferred as they are expected to perform better across a wide range of tasks.

While the last line of research has relieved the generic principle for improving zero-shot LLM prompting generalizability, these studies are largely limited to selecting among pre-existing prompts or multiple beam search samples.
Our proposed monotonic paraphrasing approach, however, can proactively rewrite a prompt or task instruction into counterparts that better satisfy the above heuristic criterion, without the need of any pre-existing candidate prompts or any training effort.

\stitle{Ensemble and Search-based Decoding}
The proposed two decoding schemes of \MODEL are connected to both ensemble decoding and search-based decoding approaches in recent studies. Ensemble decoding has been previously proposed for purposes including debiasing \cite{li-etal-2023-contrastive}, controllable generation \cite{meng2022controllable,huang-etal-2023-affective}, and cross-modality data processing \cite{liu2023prismer}.
Search-based decoding, on the other hand, is proposed to efficiently extend the search space of generation samples for satisfying specific constraints or criteria \cite{lu-etal-2022-neurologic,wuebker-etal-2012-fast,lee-berg-kirkpatrick-2022-helo,xu-etal-2023-look}.
Representative approaches include look-ahead \cite{lu-etal-2022-neurologic} and look-back \cite{xu-etal-2023-look} ones that search within the beam space towards different directions. 

The explored decoding schemes in \MODEL are inspired by both lines of studies on decoding approaches. Specifically, the first line is related to our design choice for composing the probabilities of both the paraphrasing and the target LMs in greedy generation.
And our search-based decoding scheme is specifically inspired by the look-ahead decoding strategy in the second line of studies.

\section{Conclusion}
In this paper, we propose monotonic paraphrasing (\MODEL) for automatically and iteratively generating low-perplexity prompts for given LLMs, which can lead to better task performance. Following the high-level idea of paraphrasing prompts with perplexity constraints from the target LLM, we design ensemble-based and search-based decoding strategies for efficient prompt refinement. Experiments on prompt variations of unseen tasks and instructions demonstrate the effectiveness of \MODEL in reducing prompt perplexity while enhancing task performance and model robustness. Future work may apply \MODEL on other scenarios, such as searching stealthy prompts for red-teaming.
Replacing perplexity with other criteria, such as complexity \cite{li2024famicom}, for different purposes of constrained generation is also a meaningful research direction.


\section*{Acknowledgement}

We thank the anonymous reviewers for their valuable comments.
Qin Liu was supported by a departmental fellowship.
Fei Wang was supported by the Amazon ML Fellowship.
Tianyi Yan was supported by the CURVE Fellowship.
Muhao Chen was supported by the DARPA FoundSci Grant HR00112490370, the NSF of the United States Grant ITE 2333736, and an Amazon Research Award.

\section*{Limitations}

The current investigation of \MODEL has the following limitations.
First, using the target model as the paraphrase model is a natural technical choice. However, examining other paraphrasing models, especially those customized for paraphrasing, may lead to better performance.
Second, our proposed method can not be applied to black-box LLMs such as ChatGPT\footnote{\url{https://chat.openai.com}}, PaLM \cite{chowdhery2023palm}, Claude\footnote{\url{https://www.anthropic.com/claude}}, etc.
This limitation arises from the necessity of access to the decoding phase, which in turn comes with far less computational cost due to zero training, and does not require the pre-existence of multiple prompt or instruction candidates.
Third, while we conduct experiments on a wide range of tasks, their original prompts and responses are relatively short. This actually limits the operational space of \MODEL. Experiments on tasks with longer inputs and outputs may provide additional evidence of \MODEL's effectiveness.





\bibliography{anthology,custom}
\bibliographystyle{acl_natbib}

\appendix


\newpage

\begin{center}
    {\Large\textbf{Appendices}}
\end{center}

\section{Algorithm for Search-based \MODEL}
\label{append:alg2}

\begin{algorithm}[h]
\caption{Search-based Decoding}
\label{alg:search}
\begin{algorithmic}[1]
\Require
\Statex $V$ \Comment{Vocabulary}
\Statex $P_{tar}, P_{para}$ \Comment{Target and paraphrase models}
\Statex $k$ \Comment{Number of candidate sequences to retain}
\Ensure
\Statex $x_{best}$ \Comment{Best generated sequence}

\State $X_{0} \gets \{\emptyset\}$ \Comment{Initialize with empty sequence}
\While{not $end$}
    \State $m \gets$ current decoding step
    \State $X'_{m} \gets \emptyset$
    \For{each $\mathrm{x}_{<m} \in X_{m-1}$}
        \For{each $x_m \in V$}
            \State Append $x_{<m} \circ x_m$ to $X'_{m}$
        \EndFor
    \EndFor
    \State Use $P_{para}$ for $k$ candidates for each $\mathrm{x}_{<m}$
    \State $X_{m} \gets \underset{(\mathrm{x}_{<m},x_m) \in X'_{m}}{\mathrm{\arg\,topk}} \{f(\mathrm{x}_{<m},x_m)\}$ \Comment{Select $k$ candidates with lowest perplexity}
    \State $end \gets$ If any sequence ends with $EOS$
\EndWhile
\State $x_{best} \gets$ select the sequence from $X_{m}$ with the lowest perplexity by $P_{tar}$
\State \Return $x_{best}$
\end{algorithmic}
\end{algorithm}

\section{Results on Llama-3-8B}
\label{append:llama}

Results on Llama-3-8B are illustrated in \Cref{tab:llama}.

\begin{table*}[t]
\centering
\begin{tabular}{l|l|cccc}
\toprule
Methods & Metric & GLUE Cola & Newspop & AG News & Tweet Offensive \\ \midrule
Ori. & Acc & 54.82 ± 1.73 & 73.82 ± 4.57 & 60.37 ± 7.59 & 63.25 ± 0.95 \\
 & PPL & 9.42 ± 5.08 & 17.72 ± 6.11 & 13.64 ± 4.97 & 23.64 ± 4.61 \\ \midrule
Para & Acc & 43.69 ± 9.02 & 68.45 ± 13.24 & 64.85 ± 3.73 & 64.22 ± 1.48 \\
 & PPL & 32.07 ± 8.43 & 29.41 ± 8.17 & 23.77 ± 4.32 & 46.14 ± 9.24 \\ \midrule
SPELL & Acc & 46.91 ± 2.33 & 71.67 ± 5.30 & 66.93 ± 3.70 & 64.73 ± 0.76 \\
 & PPL & 7.98 ± 1.58 & 12.47 ± 3.76 & 11.29 ± 2.79 & 8.79 ± 1.93 \\ \midrule
Mono-E & Acc & 58.35 ± 2.61 & 77.85 ± 4.01 & 65.45 ± 2.11 & 64.90 ± 0.59 \\
 & PPL & 6.64 ± 0.95 & 9.17 ± 2.59 & 8.29 ± 1.47 & 6.40 ± 3.61 \\ \bottomrule
\end{tabular}
\caption{Performance of Llama-3-8B.}
\label{tab:llama}
\end{table*}

\section{Correlation Between Accuracy and Perplexity}
\label{append:correlation}
To illustrate the interplay between task accuracy and prompt perplexity, we calculate the Pearson and Spearman correlation for each task based on the results in \Cref{tab:main}. The results are shown in \Cref{tab:correlation}.
Most tasks show a strong negative correlation between perplexity and accuracy on both models except AG News, which is also discovered by \citet{gonen-etal-2023-demystifying} that for some of the tasks, there is not a negative correlation between perplexity and accuracy. The paraphrasing may be more helpful for cases where the nuances of expression don't matter so much (or there's enough room for error), but in domains where precision of language is important, perhaps perplexity-based paraphrasing might not be as helpful.
AG News is a four-class news classification task with labels of ``World'', ``Business'', ``Sci/Tech'', and ``Sports'', which are not as easy as other binary classification tasks. As a result, the precision of the instruction for AG News is important. Additionally, we can see that the accuracy of AG News is not affected much under different methods, indicating that the accuracy is not so dependent on the perplexity of the prompt as other tasks. This indicates that \MODEL works best for tasks with high distinction among labels.

\begin{table*}[t]
\setlength\tabcolsep{2pt}
\centering
\begin{tabular}{llccccccc} \toprule
\textbf{Model} & \textbf{Metric} & \textbf{GLUE Cola} & \textbf{Newspop} & \textbf{AG News} & \textbf{IMDB} & \textbf{DBpedia} & \textbf{Emotion} & \textbf{Tweet Offensive} \\ \midrule
\multirow{2}{*}{Mistral} & Pearson & -0.62 & -0.88 & 0.19 & -0.04 & -0.70 & -0.57 & -0.48 \\ 
 & Spearman & -0.30 & -0.99 & 0.30 & -0.30 & -0.39 & -0.70 & -0.30 \\ \midrule
\multirow{2}{*}{Starling} & Pearson & -0.83 & -0.69 & 0.16 & -0.60 & -0.70 & -0.96 & -0.05 \\
 & Spearman & -0.90 & -0.49 & 0.3 & -0.3 & -0.70 & -0.89 & -0.15 \\ \bottomrule
\end{tabular}
\caption{Pearson and Spearman correlation between task accuracy and prompt perplexity w.r.t. the target model. Results are calculated based on \Cref{tab:main}.}
\label{tab:correlation}
\end{table*}

\begin{table*}[]
\setlength\tabcolsep{2pt}
\centering
\small
\begin{tabular}{lcccccccc} \toprule
\textbf{Method} & \textbf{Metric} & {\color[HTML]{2C3A4A} \textbf{GLUE Cola}} & {\color[HTML]{2C3A4A} \textbf{Newspop}} & {\color[HTML]{2C3A4A} \textbf{AG News}} & {\color[HTML]{2C3A4A} \textbf{IMDB}} & {\color[HTML]{2C3A4A} \textbf{DBpedia}} & {\color[HTML]{2C3A4A} \textbf{Emotion}} & {\color[HTML]{2C3A4A} \textbf{Tweet Offensive}} \\ \midrule
{\color[HTML]{333333} } & {\color[HTML]{333333} Acc} & {\color[HTML]{333333} 61.76 ± 2.93} & {\color[HTML]{333333} 85.28 ± 3.45} & {\color[HTML]{333333} 67.68 ± 3.47} & {\color[HTML]{333333} 84.12 ± 4.89} & {\color[HTML]{333333} 70.82 ± 2.27} & {\color[HTML]{333333} 44.32 ± 1.57} & {\color[HTML]{333333} 67.04 ± 0.95} \\
\multirow{-2}{*}{{\color[HTML]{333333} Ori.}} & {\color[HTML]{333333} PPL} & {\color[HTML]{333333} 15.72 ± 4.60} & {\color[HTML]{333333} 12.60 ± 2.40} & {\color[HTML]{333333} 12.74 ± 1.07} & {\color[HTML]{333333} 16.41 ± 0.80} & {\color[HTML]{333333} 8.10 ± 0.22} & {\color[HTML]{333333} 16.48 ± 2.14} & {\color[HTML]{333333} 19.25 ± 3.43} \\  \midrule
{\color[HTML]{333333} } & {\color[HTML]{333333} Acc} & {\color[HTML]{333333} 39.48 ± 8.19} & {\color[HTML]{333333} 76.72 ± 16.54} & {\color[HTML]{333333} 71.60 ± 2.70} & {\color[HTML]{333333} 86.60 ± 4.86} & {\color[HTML]{333333} 70.22 ± 5.71} & {\color[HTML]{333333} 43.65 ± 3.41} & {\color[HTML]{333333} 65.04 ± 1.46} \\
\multirow{-2}{*}{{\color[HTML]{333333} Para}} & {\color[HTML]{333333} PPL} & {\color[HTML]{333333} 78.01 ± 11.22} & {\color[HTML]{333333} 49.37 ± 7.81} & {\color[HTML]{333333} 28.37 ± 2.58} & {\color[HTML]{333333} 22.27 ± 0.98} & {\color[HTML]{333333} 28.34 ± 1.36} & {\color[HTML]{333333} 46.53 ± 5.30} & {\color[HTML]{333333} 53.65 ± 1.92} \\ \midrule
{\color[HTML]{333333} } & {\color[HTML]{333333} Acc} & {\color[HTML]{333333} 63.96 ± 3.88} & {\color[HTML]{333333} \textit{88.96 ± 3.04}} & {\color[HTML]{333333} \textit{72.20 ± 3.87}} & {\color[HTML]{333333} \textit{87.58 ± 3.27}} & {\color[HTML]{333333} \textit{72.07 ± 3.54}} & {\color[HTML]{2C3A4A} \textbf{47.76 ± 3.72}} & {\color[HTML]{333333} 67.00 ± 1.21} \\
\multirow{-2}{*}{{\color[HTML]{333333} Mono-E}} & {\color[HTML]{333333} PPL} & {\color[HTML]{333333} 14.89 ± 2.75} & {\color[HTML]{333333} \textit{8.33 ± 1.30}} & {\color[HTML]{333333} \textit{10.07 ± 2.93}} & {\color[HTML]{333333} \textit{13.05 ± 1.86}} & {\color[HTML]{333333} 8.85 ± 0.88} & {\color[HTML]{333333} \textit{13.67 ± 1.87}} & {\color[HTML]{333333} \textit{14.11 ± 3.53}} \\ \midrule
\multicolumn{9}{l}{{\color[HTML]{2C3A4A} \textbf{Rerank (1)}}} \\ \midrule
{\color[HTML]{333333} } & {\color[HTML]{333333} Acc} & {\color[HTML]{333333} 62.20 ± 3.47} & {\color[HTML]{333333} 86.14 ± 3.14} & {\color[HTML]{333333} 69.31 ± 4.29} & {\color[HTML]{333333} 85.02 ± 4.13} & {\color[HTML]{333333} 69.88 ± 4.33} & {\color[HTML]{333333} 47.36 ± 3.28} & {\color[HTML]{333333} 66.12 ± 2.15} \\
\multirow{-2}{*}{{\color[HTML]{333333} Budget = 10}} & {\color[HTML]{333333} PPL} & {\color[HTML]{333333} 14.27 ± 2.69} & {\color[HTML]{333333} 11.46 ± 1.99} & {\color[HTML]{333333} 11.92 ± 3.97} & {\color[HTML]{333333} 15.95 ± 2.83} & {\color[HTML]{333333} 13.15 ± 1.24} & {\color[HTML]{333333} 15.06 ± 2.71} & {\color[HTML]{333333} 15.63 ± 3.57} \\ \midrule
{\color[HTML]{333333} } & {\color[HTML]{333333} Acc} & {\color[HTML]{333333} 63.84 ± 2.46} & {\color[HTML]{333333} 85.86 ± 5.36} & {\color[HTML]{333333} 68.28 ± 5.16} & {\color[HTML]{333333} 86.52 ± 5.86} & {\color[HTML]{333333} 69.64 ± 3.53} & {\color[HTML]{333333} 47.60 ± 5.25} & {\color[HTML]{333333} 67.74 ± 1.45} \\
\multirow{-2}{*}{{\color[HTML]{333333} Budget = 50}} & {\color[HTML]{333333} PPL} & {\color[HTML]{333333} 13.54 ± 3.16} & {\color[HTML]{333333} 10.53 ± 2.85} & {\color[HTML]{333333} 10.83 ± 3.24} & {\color[HTML]{333333} 13.44 ± 2.22} & {\color[HTML]{333333} 10.31 ± 1.85} & {\color[HTML]{333333} 14.14 ± 3.46} & {\color[HTML]{333333} 13.94 ± 3.86} \\ \midrule
{\color[HTML]{333333} } & {\color[HTML]{333333} Acc} & {\color[HTML]{2C3A4A} \textbf{64.58 ± 3.50}} & {\color[HTML]{2C3A4A} \textbf{89.82 ± 2.04}} & {\color[HTML]{2C3A4A} \textbf{72.33 ± 4.17}} & {\color[HTML]{2C3A4A} \textbf{87.61 ± 4.93}} & {\color[HTML]{2C3A4A} \textbf{72.52 ± 3.16}} & {\color[HTML]{333333} \textit{47.70 ± 3.56}} & {\color[HTML]{2C3A4A} \textbf{68.5 ± 1.84}} \\
\multirow{-2}{*}{{\color[HTML]{333333} Budget = 100}} & {\color[HTML]{333333} PPL} & {\color[HTML]{2C3A4A} \textbf{10.22 ± 2.32}} & {\color[HTML]{2C3A4A} \textbf{7.53 ± 1.82}} & {\color[HTML]{2C3A4A} \textbf{9.52 ± 4.20}} & {\color[HTML]{2C3A4A} \textbf{12.88 ± 3.01}} & {\color[HTML]{2C3A4A} \textbf{8.16 ± 1.67}} & {\color[HTML]{2C3A4A} \textbf{13.05 ± 1.34}} & {\color[HTML]{2C3A4A} \textbf{11.56 ± 4.44}} \\ \midrule
\multicolumn{9}{l}{{\color[HTML]{2C3A4A} \textbf{Rerank (2)}}} \\ \midrule
{\color[HTML]{333333} } & {\color[HTML]{333333} Acc} & {\color[HTML]{333333} 62.35 ± 4.77} & {\color[HTML]{333333} 86.96 ± 4.61} & {\color[HTML]{333333} 70.02 ± 5.27} & {\color[HTML]{333333} 84.86 ± 4.71} & {\color[HTML]{333333} 69.31 ± 3.94} & {\color[HTML]{333333} 46.10 ± 7.29} & {\color[HTML]{333333} 66.63 ± 4.53} \\
\multirow{-2}{*}{{\color[HTML]{333333} Budget = 10}} & {\color[HTML]{333333} PPL} & {\color[HTML]{333333} 15.63 ± 3.85} & {\color[HTML]{333333} 12.27 ± 3.99} & {\color[HTML]{333333} 13.47 ± 3.63} & {\color[HTML]{333333} 16.43 ± 2.17} & {\color[HTML]{333333} 15.37 ± 2.14} & {\color[HTML]{333333} 16.64 ± 3.88} & {\color[HTML]{333333} 18.87 ± 4.48} \\ \midrule
{\color[HTML]{333333} } & {\color[HTML]{333333} Acc} & {\color[HTML]{333333} 62.08 ± 3.97} & {\color[HTML]{333333} 86.50 ± 5.10} & {\color[HTML]{333333} 71.12 ± 6.97} & {\color[HTML]{333333} 86.03 ± 5.24} & {\color[HTML]{333333} 70.13 ± 4.23} & {\color[HTML]{333333} 46.90 ± 5.83} & {\color[HTML]{333333} 65.15 ± 5.03} \\
\multirow{-2}{*}{{\color[HTML]{333333} Budget = 50}} & {\color[HTML]{333333} PPL} & {\color[HTML]{333333} 14.42 ± 4.16} & {\color[HTML]{333333} 10.24 ± 4.36} & {\color[HTML]{333333} 11.69 ± 2.96} & {\color[HTML]{333333} 14.33 ± 3.55} & {\color[HTML]{333333} 13.62 ± 2.77} & {\color[HTML]{333333} 15.21 ± 3.38} & {\color[HTML]{333333} 18.53 ± 4.54} \\ \midrule
{\color[HTML]{333333} } & {\color[HTML]{333333} Acc} & {\color[HTML]{333333} \textit{64.35 ± 4.91}} & {\color[HTML]{333333} 87.82 ± 4.48} & {\color[HTML]{333333} 71.72 ± 6.07} & {\color[HTML]{333333} 86.64 ± 5.16} & {\color[HTML]{333333} 71.98 ± 3.85} & {\color[HTML]{333333} 47.18 ± 4.63} & {\color[HTML]{333333} \textit{67.80 ± 8.95}} \\
\multirow{-2}{*}{{\color[HTML]{333333} Budget = 100}} & {\color[HTML]{333333} PPL} & {\color[HTML]{333333} \textit{11.67 ± 4.37}} & {\color[HTML]{333333} 8.93 ± 3.15} & {\color[HTML]{333333} 10.61 ± 4.89} & {\color[HTML]{333333} 13.47 ± 3.84} & {\color[HTML]{333333} \textit{8.36 ± 3.57}} & {\color[HTML]{333333} 14.41 ± 2.95} & {\color[HTML]{333333} 16.65 ± 3.53} \\ \bottomrule
\end{tabular}
\caption{Reranking for ensemble-based \MODEL. The best results of each task are \textbf{bolded} and the second-best are \emph{italicized}.}
\label{tab:reranking}
\end{table*}
\begin{table}[]
\centering
\setlength\tabcolsep{2pt}
\small
\begin{tabular}{lccccc} \toprule
\textbf{} & \textbf{Newspop} & \textbf{IMDB} & \textbf{DBpedia} & \textbf{Emotion} & \textbf{Tweet} \\ \midrule
Ori & 0.949 & 0.952 & 0.895 & 0.839 & 0.974 \\
Para & 0.813 & 0.951 & 0.835 & 0.645 & 0.871 \\
Mono-E & 0.956 & 0.895 & 0.927 & 0.931 & 0.968 \\ \bottomrule
\end{tabular}
\caption{Pearson correlation between the perplexity of standalone prompt and its instantiated prompts. Pearson correlation is high (close to $1.0$) in most
cases, indicating a high correlation.}
\label{tab:gap}
\end{table}
\begin{table}[]
\centering
\setlength\tabcolsep{2pt}
\small
\begin{tabular}{lccc} \toprule
\textbf{} & \textbf{Ori} & \textbf{Para} & \textbf{Mono-E} \\ \midrule
Un. & 56.67 & 58.31 & 65.94 \\
Cha. & 51.03 (-9.95\%) & 53.62 (-8.04\%) & 65.81 (-0.20\%) \\
Word & 52.41 (-7.51\%) & 57.39 (-1.58\%) & 67.23 (+1.96\%) \\
Sem. & 49.42 (-12.79\%) & 54.22 (-7.01\%) & 64.35 (-2.4\%) \\ \bottomrule
\end{tabular}
\caption{Performance degradation of Alpaca model compared with the unperturbed prompt. \emph{Un.}, \emph{Cha.}, and \emph{Sem.} are short for unperturbed, character, and semantic, respectively, referring to different prompt perturbation methods.}
\label{tab:task2}
\end{table}
\begin{table*}[t]
\setlength\tabcolsep{3pt}
\renewcommand\arraystretch{1.2}
\centering
\small
{\footnotesize
\begin{tabularx}{\linewidth}{lXc}
\toprule
 &  Content & BertScore  \\ \midrule
Ori Prompt & In this task, you are presented with a short article. Your objective is to classify the article according to its category using the following labels: World, Sports, Business, Sci/Tech. Tag the text as  ``World''  if it involves information about global events or issues. Mark it as ``Sports'' if it deals with sports-related content. Label it ``Business'' if it pertains to business affairs, markets, or economics. Assign ``Sci/Tech'' if it covers scientific or technological developments. & -- \\ \midrule
Para & In this \textbf{assignment}, you are given a brief article. Your goal is to \textbf{categorize} the article based on its topic using these tags: World, Sports, Business, Sci/Tech. Identify it as ``World'' if it discusses \textbf{international} events or matters. Designate it as ``Sports'' if it focuses on athletic content. Categorize it as ``Business'' if it concerns commercial activities, markets, or the economy. Label it ``Sci/Tech'' if it reports on advancements in science or technology. & 86.02  \\
\midrule
Mono-E & In your \textbf{task} you will have an input which consists of a single article. You have to sort this article according to its category using a list which consists of 4 categories: ``World'', ``Sci/tech'', ``Business'', or ``Sports''. You should mark it with the appropriate label based on the content. The article should be labeled with the ``Sports'' category, when its topic deals with sports. When the article is about business, you need to mark it as a 'business' label. If you are talking about science and tech developments then label the content with ``Sci/Tech'', if it is a \textbf{world-related} issue or event then label it with ``World''. & 89.63  \\ \bottomrule
\end{tabularx}}
\caption{Case study on the paraphrase results and BERTScore. Para tends to use new vocabulary for diversity, while Mono-E prefers words that are more familiar to the target model, which is often the same as in the original prompt.}
\label{tab:case_study}
\end{table*}
\begin{table*}[ht]
\setlength\tabcolsep{3pt}
\centering
\small
\begin{tabular}{lcccccccc} \toprule
\textbf{Method} & \textbf{Metric} & \textbf{GLUE Cola} & \textbf{Newspop} & \textbf{AG News} & \textbf{IMDB} & \textbf{DBpedia} & \textbf{Emotion} & \textbf{Tweet} \\ \midrule
\multirow{2}{*}{Ori.} & Acc & 61.76 ± 2.93 & 85.28 ± 3.45 & 67.68 ± 3.47 & 84.12 ± 4.89 & 70.82 ± 2.27 & 44.32 ± 1.57 & 67.04 ± 0.95 \\
 & PPL & 15.72 ± 4.60 & 12.60 ± 2.40 & 12.74 ± 1.07 & 16.41 ± 0.80 & 8.10 ± 0.22 & 16.48 ± 2.14 & 19.25 ± 3.43 \\ \midrule
\multirow{2}{*}{Para} & Acc & 39.48 ± 8.19 & 76.72 ± 16.54 & 71.60 ± 2.70 & 86.60 ± 4.86 & 70.22 ± 5.71 & 43.65 ± 3.41 & 65.04 ± 1.46 \\
 & PPL & 78.01 ± 11.22 & 49.37 ± 7.81 & 28.37 ± 2.58 & 22.27 ± 0.98 & 28.34 ± 1.36 & 46.53 ± 5.30 & 53.65 ± 1.92 \\ \midrule
\multirow{2}{*}{Mono-E} & Acc & \textit{63.96 ± 3.88} & 88.96 ± 3.04 & \textit{72.20 ± 3.87} & 87.58 ± 3.27 & \textit{72.07 ± 3.54} & \textbf{47.76 ± 3.72} & 67.00 ± 1.21 \\
 & PPL & 14.89 ± 2.75 & 8.33 ± 1.30 & 10.07 ± 2.93 & 13.05 ± 1.86 & 8.85 ± 0.88 & 13.67 ± 1.87 & 14.11 ± 3.53 \\ \midrule
\multirow{2}{*}{$\alpha$ = 0.1} & Acc & 60.80 ± 4.28 & 84.98 ± 3.41 & 69.09 ± 4.01 & 85.52 ± 1.72 & 70.42 ± 3.66 & 44.80 ± 2.31 & 65.56 ± 5.07 \\
 & PPL & 15.61 ± 2.51 & 12.32 ± 1.01 & 12.03 ± 2.96 & 15.70 ± 1.85 & 9.61 ± 0.61 & 16.16 ± 1.11 & 18.49 ± 2.84 \\ \midrule
\multirow{2}{*}{$\alpha$ = 0.2} & Acc & 61.80 ± 5.63 & 85.32 ± 3.56 & 71.49 ± 4.32 & 85.08 ± 4.97 & 70.54 ± 5.23 & 44.04 ± 2.44 & 66.80 ± 5.19 \\
 & PPL & 15.22 ± 2.06 & 11.56 ± 1.27 & 11.68 ± 3.25 & 15.30 ± 1.55 & 9.23 ± 0.79 & 15.71 ± 0.20 & 16.92 ± 2.68 \\ \midrule
\multirow{2}{*}{$\alpha$ = 0.3} & Acc & 61.54 ± 5.85 & 86.72 ± 4.94 & \textbf{72.4 ± 2.77} & 86.68 ± 4.21 & 71.90 ± 3.91 & 44.96 ± 2.67 & \textit{67.56 ± 3.59} \\
 & PPL & 15.18 ± 2.84 & 10.21 ± 1.22 & 11.06 ± 2.74 & 14.42 ± 1.05 & 8.95 ± 0.75 & 15.21 ± 1.04 & 15.87 ± 3.61 \\ \midrule
\multirow{2}{*}{$\alpha$ = 0.4} & Acc & 62.72 ± 4.37 & \textit{89.44 ± 3.08} & 71.31 ± 2.27 & \textbf{87.86 ± 3.37} & \textbf{72.18 ± 3.44} & 45.58 ± 1.55 & \textbf{67.70 ± 1.60} \\ 
 & PPL & 14.93 ± 2.91 & 9.16 ± 0.57 & 10.39 ± 3.56 & 13.81 ± 2.60 & 9.04 ± 1.18 & 14.35 ± 0.90 & 15.18 ± 3.72 \\ \midrule
\multirow{2}{*}{$\alpha$ = 0.6} & Acc & \textbf{64.30 ± 6.63} & \textbf{90.86 ± 3.24} & 71.66 ± 10.09 & \textit{87.74 ± 4.04} & 71.36 ± 6.25 & \textit{47.12 ± 1.66} & 66.26 ± 3.24 \\
 & PPL & 11.49 ± 1.00 & 7.53 ± 0.68 & 9.38 ± 0.88 & 12.15 ± 2.47 & 7.11 ± 1.08 & 13.12 ± 0.79 & 13.29 ± 2.58 \\ \midrule
\multirow{2}{*}{$\alpha$ = 0.7} & Acc & 41.50 ± 13.68 & 54.52 ± 22.29 & 50.56 ± 21.95 & 64.84 ± 16.01 & 48.12 ± 16.26 & 38.62 ± 1.59 & 59.34 ± 7.75 \\
 & PPL & 8.80 ± 0.42 & 3.84 ± 0.91 & 3.43 ± 0.36 & 4.99 ± 0.28 & 2.34 ± 0.25 & 4.28 ± 0.55 & 6.58 ± 1.40 \\ \bottomrule
\end{tabular}
\caption{Results with varying $\alpha$ values for Mono-E on Mistral-7B. The best accuracy for each task is in \textbf{bold}. For $\alpha$ = $0.4$ or $0.6$, the model shows outstanding performance on most of the datasets. But increasing alpha to $0.7$ results in a nonsensical prompt though the perplexity is super low, leading to corrupted task performance.}
\label{tab:alpha}
\end{table*}

\section{Reranking for \MODEL}
\label{sec:reranking}

\subsection{Implementation Setting}
We implemented reranking on top of the linear combination of $P_{tar}$ and $P_{para}$ under two settings: (1) the coefficient $\alpha$ equals $0.5$, which is kept consistent with the results presented in \Cref{tab:main} for direct comparison; (2) the coefficient is randomly selected from $[0.1, 0.2, …, 0.6]$ for a larger search space. In both cases, we allow sampling from the top $3$ tokens at every decoding step for randomness.

For each task, we use the $5$ prompts as the original prompts (the same setting as \Cref{tab:main}) and randomly generate $100$ paraphrases with the linear combination of $P_{tar}$ and $P_{para}$ for every single original prompt. For the generated $100$ paraphrases, we use the same scheme as is described in \Cref{sec:exp_setting} to calculate the perplexity: each paraphrase is instantiated with $1,000$ random test samples, and the perplexity is averaged across all the $1,000$ instantiated prompts. For budget $k$, we select the first $k$ paraphrases from the $100$ candidates and choose the prompt with the lowest average instantiated perplexity to be tested on the task.

\subsection{Analysis}
For the reranking setting (1), with additional randomness when sampling paraphrases from the linear combination of $P_{tar}$ and $P_{para}$, a large enough budget (i.e., $100$ paraphrases) can consistently catch paraphrased prompts with lower perplexity which achieve better task performance (\Cref{tab:reranking}). However, with a small budget of $10$ or even $50$, reranking can hardly surpass Mono-E in either perplexity or task performance. The reason is that with randomness allowed, the paraphrase is no longer decoded greedily at each token, which undermines the constraint on the perplexity and leads to a large variation of the perplexity of the resulting paraphrase. This further affects the performance of these paraphrased prompts, which in turn proves the correlation between perplexity and task accuracy. Generally, we can expect vanilla Mono-E and reranking to break even with a large budget of around $50$ paraphrases.

The reranking setting (2) is not as effective as setting (1), where even a budget of $100$ could not surpass the vanilla Mono-E. The reason is that although a large search space is allowed with randomness on the combination coefficient $\alpha$, the constraint on perplexity while decoding is undermined even further. With more randomness for paraphrase sampling, the results of setting (2) also show a higher standard deviation than setting (1).

\section{Correlation Between Standalone and Instantiated Prompts}
\label{sec:gap}
To bridge the intuition gap, we investigate the correlation between the standalone prompt and its instantiated prompts. We instantiate each task-specific prompt with $1,000$ input instances and the perplexity used for correlation evaluation is averaged over $5$ prompts for each task. As shown in \Cref{tab:gap}, the Pearson correlation is high (close to 1.0) in most cases for different datasets despite slight noise for Emotion under greedy paraphrase of prompts. We can conclude that there is a high correlation between instantiated and standalone prompts. Thus, it is valid to use the perplexity of the standalone prompt optimized by $P_{tar}$ to serve as the constraint.

\section{Unperturbed Accuracy for Task II}
\label{append:taskII}
\Cref{tab:task2} shows the task performance of three perturbation methods (Alpaca’s average accuracy across $4$ GLUE datasets) and performance degradation compared with the unperturbed prompt (as shown in the brackets). We can witness a narrow performance gap between perturbed and unperturbed original prompts for Mono-E, which is more robust than directly using the given prompt (Ori) or the greedily paraphrased one (Para). The reason is that Mono-E largely mitigates the effect of adversarial perturbations on the prompts and successfully rewrites the given prompts (whether perturbed or unperturbed) into higher-quality versions.

\section{BERTScore of Para and \MODEL}
\label{append:bertscore}
We provide a case study on the paraphrase results and the according BERTScore in \Cref{tab:case_study}. We could see that both Para and Mono-E largely preserve the semantic meaning of the original prompt. Further, Para tends to use new vocabulary for diversity, while Mono-E prefers words that are more familiar to the target model, which is often the same as in the original prompt. This might be the reason why Para results in lower BertScore compared to Mono-E and Mono-S.

\section{Ablation Study on Alpha Value}
\label{sec:alpha}

We provide results with various alpha values in \Cref{tab:alpha}, which illustrates the impact of alpha values on model accuracy and paraphrase perplexity across various datasets on the Mistral-7B model. With the increase of $\alpha$ value, the perplexity (PPL) of the paraphrase monotonically goes down due to the design and motivation of our method. Given the low perplexity, we compare the task accuracy of various alpha values.
For $\alpha$ = $0.4$ or $0.6$, the model shows outstanding performance on some of the datasets. But increasing alpha to $0.7$ results in a nonsensical prompt though the perplexity is super low. From the results we can also see that the performance just corrupted with this nonsensical prompt. The setting of $\alpha$ should seek a balance between lower perplexity and the preservation of the original semantics. As a result, alpha = $0.5$ (Mono-E) offers a balanced and reliable performance, making it a suitable default choice.

\end{document}